\documentclass[nohyperref]{article}

\usepackage{microtype}
\usepackage{graphicx}
\usepackage{subfigure}
\usepackage{booktabs} 

\usepackage{enumitem}
\usepackage{multirow}
\usepackage{tabularx}
\usepackage[ruled]{algorithm2e}
\usepackage{amsfonts}
\usepackage{amssymb}
\usepackage{amsxtra}
\usepackage{diagbox}
\usepackage{graphicx}
\usepackage{subfigure}
\usepackage{amsmath}
\usepackage{caption}
\usepackage{threeparttable}
\usepackage{wrapfig}
\usepackage{enumitem}

\usepackage{hyperref}

\usepackage[accepted]{icml2023}

\usepackage{amsmath}
\usepackage{amssymb}
\usepackage{mathtools}
\usepackage{amsthm}
\usepackage{color}
\newcommand{\tabincell}[2]{\begin{tabular}{@{}#1@{}}#2\end{tabular}}
\usepackage[capitalize,noabbrev]{cleveref}

\theoremstyle{plain}

\theoremstyle{definition}

\theoremstyle{remark}

\usepackage[textsize=tiny]{todonotes}

\icmltitlerunning{Disentangled Representation Learning with Large Language Models for Text-Attributed Graphs}

\begin{document}

\twocolumn[
\icmltitle{Disentangled Representation Learning with Large Language Models for Text-Attributed Graphs}




\begin{icmlauthorlist}
\icmlauthor{Yijian Qin}{ts}
\icmlauthor{Xin Wang}{ts}
\icmlauthor{Ziwei Zhang}{ts}
\icmlauthor{Wenwu Zhu}{ts}
\end{icmlauthorlist}

\icmlaffiliation{ts}{Department of Computer Science and Technology, Tsinghua University}

\icmlcorrespondingauthor{Wenwu Zhu}{wwzhu@tsinghua.edu.cn}
\icmlcorrespondingauthor{Xin Wang}{xin\_wang@tsinghua.edu.cn}
\icmlkeywords{Machine Learning, ICML}

\vskip 0.3in
]



\printAffiliationsAndNotice{}  


\begin{abstract}
Text-attributed graphs (TAGs) are prevalent on the web and research over TAGs such as citation networks, e-commerce networks and social networks has attracted considerable attention in the web community. Recently, large language models (LLMs) have demonstrated exceptional capabilities across a wide range of tasks. However, the existing works focus on harnessing the potential of LLMs solely rely on prompts to convey graph structure information to LLMs, thus suffering from insufficient understanding of the complex structural relationships within TAGs. To address this problem, in this paper we present the Disentangled Graph-Text Learner (DGTL) model, which is able to enhance the reasoning and predicting capabilities of LLMs for TAGs. Our proposed DGTL model incorporates graph structure information through tailored disentangled graph neural network layers, enabling LLMs to capture the intricate relationships hidden in text-attributed graphs from multiple structural factors. Furthermore, DGTL operates with frozen pre-trained LLMs, reducing computational costs and allowing much more flexibility in combining with different LLM models. Experimental evaluations demonstrate the effectiveness of the proposed DGTL model on achieving superior or comparable performance over state-of-the-art baselines. Additionally, we also demonstrate that our DGTL model can offer natural language explanations for predictions, thereby significantly enhancing model interpretability.
\end{abstract}

\section{Introduction}
\label{submission}

Text-attributed graphs (TAGs) are employed to represent a group of structured data where textual entities are connected by graph relations. TAGs are ubiquitous on the web, such as citation networks, e-commerce networks, social media, recommendation systems, web pages etc. As such, representation learning on TAGs has become an important research problem recently in the web community, where the TAGs are explored to capture rich semantic relationships and dependencies among connected textual elements, providing valuable contexts for better understanding and reasoning in various downstream tasks. Classical text-attributed graph representation approaches normally utilize graph neural networks (GNNs) to capture structural information, transforming textual attributes into shallow or hand-crafted representations such as bag-of-words or skip-gram features, which will then be used for prediction tasks in TAG~\cite{gcn}. Some works also use natural language processing models to enhance GNN classifiers by augmenting the node features and capture the rich semantic information~\cite{zhang2020graph,yang2021bert}. 

Recent advancements in machine learning and artificial intelligence have witnessed the emergence of large language models (LLMs) that exhibit unprecedented capabilities in various tasks~\cite{zhao2023survey}. These models have demonstrated remarkable proficiency in natural language processing related tasks including language generation, machine translation and sentiment analysis, as well as other fields such as recommendation system~\cite{wu2023survey}, social network analysis~\cite{gao2023s}, code analysis~\cite{chen2021evaluating}, bioinformatics~\cite{thirunavukarasu2023large} and many more. 

\begin{figure*}
  \centering
  \includegraphics[width=\linewidth]{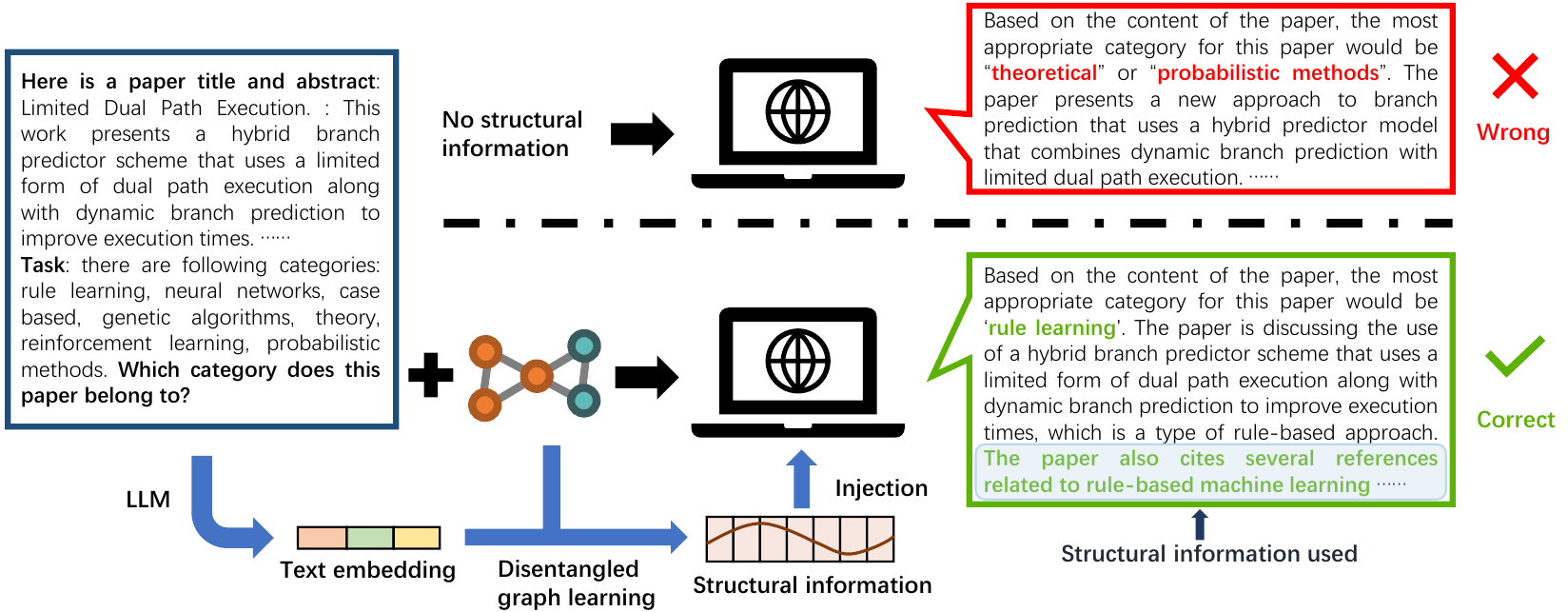}
  \caption{An illustration of using LLMs to solve paper classification task on TAGs. Top: existing methods fail to give correct answer because of lacking structural information on the TAG. Bottom: our method can predict the correct label by utilizing the structural information, i.e., the reference information in the citation graph.}
\end{figure*}

Therefore, the advent of LLMs has promoted the direct exploration of LLMs to solve prediction problems in TAG without GNN classifiers. These existing approaches solely rely on prompts to convey graph structure information to LLMs, suffering from insufficient understanding of the complex structural relationships within TAGs. For example, Graph-LLM~\cite{chen2021evaluating} uses neighbor summary to generate prompts with structural information, while InstructGLM~\cite{ye2023natural} directly describes all the neighbors in the prompt. Nevertheless, only utilizing prompts to pass information about graph structure over LLMs hinders the model's ability to capture and utilize the intricate relationships and dependencies encoded in the graph structures of TAGs, resulting in limited capability of exploiting the potential power of LLMs in TAG tasks. 

To tackle this problem, in this paper we go beyond the vanilla prompt-based methods and effectively integrate graph structure information into LLMs, in order to enable the holistic utilization of LLMs' exceptional powers in TAG tasks. However, achieving this goal poses the following technical challenges. First, TAGs usually contain rich yet entangled structural information, i.e., not all the information is relevant or helpful for downstream tasks. LLMs need to effectively filter out and extract the pertinent information while disregarding the irrelevant details carried in the graph structure. Second, adapting LLMs to a specific TAG task is challenging, given that LLMs typically require extensive training and learning of task-specific knowledge. The process of fine-tuning LLMs for particular tasks involves striking the balance between maintaining the pre-trained knowledge and acquiring new knowledge specific to the target tasks. 

To solve these challenges, we propose the Disentangled Graph Text Learner (DGTL) model in this paper, which can boost LLMs in deeply exploiting structural information to enhance their reasoning and predicting capabilities for TAG tasks. The proposed DGTL model first encodes raw text information carried in TAGs using a pre-trained LLM. Then, a group of tailored disentangled GNN layers is developed to capture graph neighborhood information in TAGs from multiple structural factors. By injecting the learned features with disentangled factors into the LLM predictor, we enable the model to comprehend complex graph structure information in TAGs. Moreover, DGTL allows the pre-trained LLMs to remain frozen, thereby reducing computation costs and mitigating the risk of catastrophic forgetting in LLMs. This flexibility enables DGTL to be compatible with different LLM models, ensuring its practical applicability. Overall, DGTL is able to serve as a general framework for combining text-attributed graph learning and natural language modeling to improve the explainability and predictive performance of LLMs for TAG tasks.

To demonstrate the effectiveness of our proposed method, we compare it with state-of-the-art approaches on various TAG benchmarks. Our method achieves superior or comparable results with baseline methods. Additionally, we demonstrate that DGTL can offer human-understandable explanations for the model predictions in natural language. 

In summary, we make the following contributions:
\begin{itemize}[leftmargin = *]
    \item We propose Disentangled Graph Text Learner (DGTL), a novel model which deeply exploits graph structure information to enhance the reasoning and predicting capabilities of LLMs for TAG tasks. DGTL also serves as a general framework for integrating structural analysis abilities of GNNs with the powerful language modeling capabilities of LLMs. 
    \item We propose tailored disentangled GNN layers to capture graph neighborhood information from multiple structural factors, enabling the LLMs to comprehend complex graph structure information in TAGs. 
    \item Our proposed DGTL enables pre-trained LLMs to remain frozen, benefiting from reduced computation costs as well as mitigating the risk of catastrophic forgetting. 
    \item We conduct extensive experiments on various TAG benchmarks and compare DGTL with state-of-the-art baselines. The results demonstrate that DGTL is able to achieve superior or comparable performance, as well as provide users with human understandable natural language explanations for the model's predictions.
\end{itemize}

\section{Related Works}\label{sec:related}
\subsection{Text-attributed Graphs}
\label{sec:tag}
Text-attributed graphs (TAGs) have gained significant attention in the field of graph machine learning in recent years~\cite{wang2017community,wang2019heterogeneous,huang2023prompt,duan2023simteg}. A TAG is a type of graphs where each node is associated with a text attribute. This representation captures the rich semantic relationships and dependencies among textual elements, making TAGs valuable for understanding and reasoning tasks. Commonly used TAG benchmark datasets include Cora, CiteSeer, PubMed~\cite{sen2008collective}, and OGBN-arXiv~\cite{ogb}, where nodes represent papers and edges represent reference relationships.

Message-passing GNNs~\cite{gcn, gat, gin, graphsage} have been proposed as an effective framework for graph machine learning following the neighborhood aggregation scheme. At each layer, nodes learn representations by aggregating their neighbors' representations~\cite{gilmer2017neural}. GNNs have also made significant progress in TAG tasks by considering both node attributes and graph structures. Classical GNN pipelines typically handle text attributes by converting them into shallow or hand-crafted features such as bag-of-words~\cite{zhang2010understanding} or skip-gram~\cite{mikolov2013distributed} representations. However, with the advancements in natural language processing, there has been a shift towards utilizing language models to generate more comprehensive node features based on the text attribute. This approach allows for a deeper understanding and representation of learning on TAGs.

Despite the progress made by LLMs and GNNs in capturing textual or structural information for representation learning on TAGs, there are still large rooms for improvement. The integration of these models can lead to enhanced performance and more effective utilization of the rich information contained within TAGs.

\subsection{LLMs for Graph Tasks}
Recent researches have also delved into the exploration of leveraging LLMs directly for addressing graph-related tasks~\cite{zhang2023large}. The fundamental concept behind this approach is to convert graph data, including both structural components and features, as well as graph tasks, into natural language representations. By treating graph problems as conventional NLP problems, researchers have unlocked the potential of utilizing LLMs for graph-related tasks. In the subsequent sections, we present a comprehensive overview of these recent advancements in the field.

Pioneer researches begin with explorations on synthetic graph tasks. NLGraph~\cite{wang2023can} reorganizes graphs as natural language description and conducts a systematic evaluation of LLMs on eight graph reasoning tasks in natural language, including connectivity, shortest path, maximum flow, simulating GNNs, etc. Meanwhile, GPT4Graph~\cite{guo2023gpt4graph} also conducts extensive experiments by converting graphs into specific code formats. They evaluate the graph understanding capabilities of LLMs across ten distinct tasks, including structure understanding tasks and semantic understanding tasks. LLMtoGraph~\cite{liu2023evaluating} also tests GPT-3.5 and GPT-4 for various graph tasks by natural language description and makes some interesting observations. 

More recently, several works carry out explorations on TAGs with LLMs. Graph-LLM~\cite{chen2023exploring} systematically investigates two strategies on TAGs: LLMs-as-Enhancers and LLMs-as-Predictors. The former strategy uses LLM to enhance the representations of text attributes of nodes before passing them to GNNs, while the latter one directly employs LLM as TAG task predictors with natural language prompts. They also explore using ego-graph description and neighbor summary to incorporate structural information by prompt. InstructGLM~\cite{ye2023natural} expands the vocabulary by creating new tokens for every node in the TAG, which enables tuning LLMs to handle various TAG tasks in a generative manner. Nevertheless, the existing methods use prompt to convey neighborhood information for downstream LLMs, which faces several challenges, such as the issue of excessive neighboring information leading to lengthy prompt texts. GraphQA~\cite{he2024g} is a question-answering benchmark on multiple applications including scene graph understanding, common sense reasoning, and knowledge graph reasoning.

In our method, we employ a disentangled graph learning approach to compress the neighboring information on the TAG into a small number of tokens. This enables LLMs to learn and utilize the rich knowledge contained within these compressed tokens for downstream inference tasks. Besides, our method follows a delta-tuning scheme where the LLM is kept frozen and only a small number of parameters are tuned~\cite{ding2023parameter}, making our method easy to integrate with off-the-shelf LLMs.

\section{Problem Formulation and Preliminaries}\label{sec:formulation}
\subsection{Text-attributed Graphs}
A text-attributed graph can by formulated as $\mathcal{G}=(\mathcal{V},\mathbf{A},\mathbf{y})$, where $\mathcal{V}$ denotes the set of nodes, $\mathbf{A}$ denotes the adjacent matrix, and $\mathbf{y}$ denotes the labels of nodes. In TAGs, each node is associated with a text description, e.g., the abstract of papers for citation graphs. Before the message-passing procedure in the GNNs, we need to process the raw texts into real-valued features, i.e., text embeddings.

In this paper, we focus on node classification, one of the most typical tasks on TAGs. We adopt the semi-supervised settings, where the text information of all the node set $\mathcal{V}$ and $\mathbf{A}$ is given at the training procedure, as well as a part of the node labels $\{\mathbf{y}_u|u\in \mathcal{V}_{tr}\}$, where $\mathcal{V}_{tr}$ is the  training node set. The task aims at predicting the labels $\{\mathbf{y}_u|u\in \mathcal{V}_{te}\}$ of the testing node set $\mathcal{V}_{te}$.

\subsection{Graph Neural Network}
GNNs are state-of-the-art models for graph machine learning, which typically follow a message passing scheme where nodes aggregate information
from their neighbors in each layer formulated as:
\begin{align}\label{eq:mpnn}
    \mathbf{m}_i^{(l)} &= \text{Agg}(\mathbf{h}_j^{(l)}|j\in \mathcal{N}_i),\\
    \label{equ:t}
    \mathbf{h}_i^{(l+1)} &= \text{Update}(\mathbf{m}_i^{(l)}),
\end{align}
where $\mathbf{h}_i^{(l)}$ is the representation of node $i$ at the $l$-th layer, $\mathcal{N}_i$ denotes the neighbors of node $i$ derived from the adjacent matrix, $\text{Agg}(\cdot)$ is the aggregation function, $\text{Update}(\cdot)$ is an updating function between two node representations.

\begin{figure*}
  \centering
  \includegraphics[width=\linewidth]{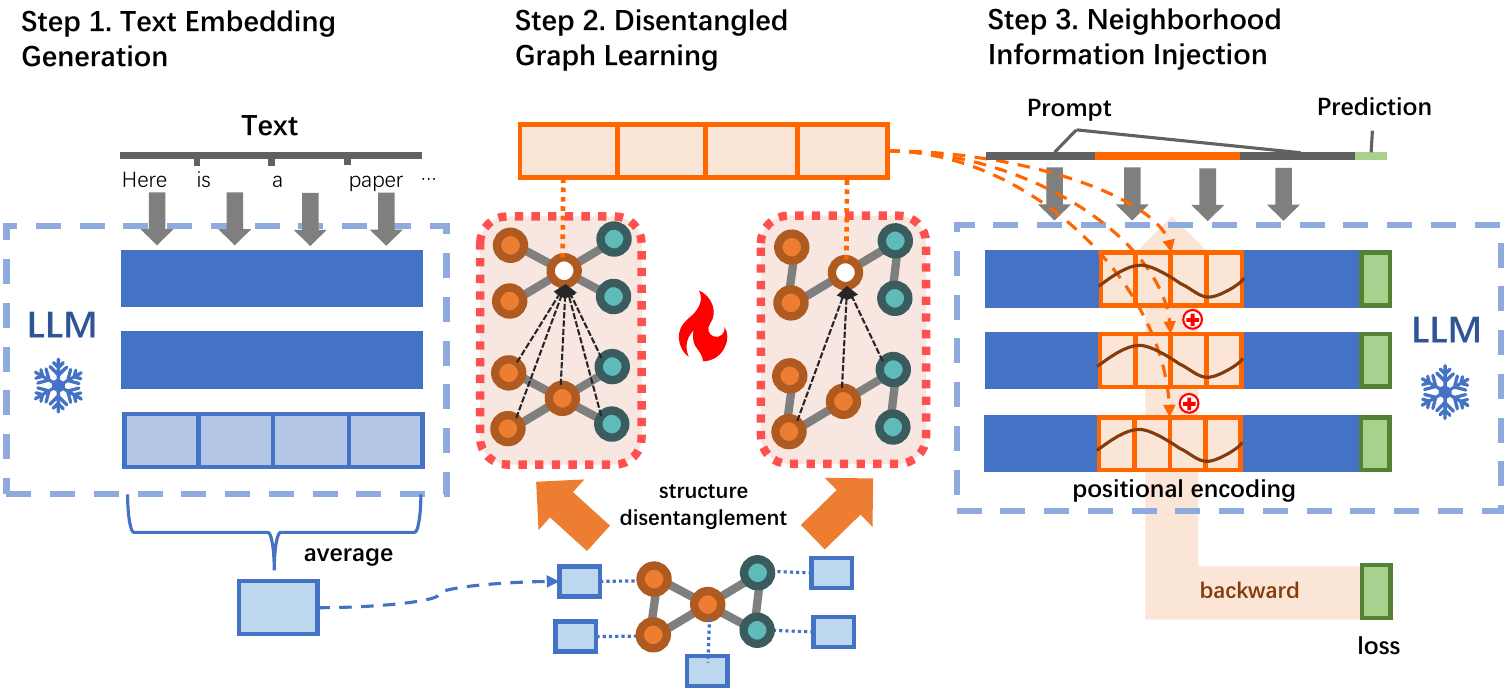}
  \caption{An overview of our proposed DGTL method. \textbf{Step 1:} Generating text embedding by taking average of the features at the last layer in the upstream LLM. \textbf{Step 2:} Using our proposed disentangled graph learning to learn embeddings with diverse structural information. \textbf{Step 3:} Injecting the features with neighborhood information into the downstream LLM.}
  \label{fig:framework}
\end{figure*}

\subsection{Large Language Models}
Large language models (LLMs) have revolutionized natural language processing tasks by demonstrating remarkable capabilities in understanding and generating human-like text. One of the key architectural advancements that underpins the success of LLMs is the transformer model, which has become the de facto standard for modeling sequential data and computer vision.

The transformer architecture~\cite{vaswani2017attention} is based on the principle of self-attention. The attention mechanism enables the model to weigh the importance of different parts of the input sequence when making predictions. The attention mechanism calculates attention scores between each pair of positions in the input sequence, allowing the model to focus on relevant information while disregarding irrelevant or redundant elements. This attention mechanism is crucial for capturing long-range dependencies and contextual relationships in text. The self-attention in current decoder-based LLMs can be formulated as follows:
\begin{align}\label{eq:att}
    \text{Attention}(\mathbf{H}) = \text{softmax}(\frac{f_q(\mathbf{H})f_k(\mathbf{H})}{\sqrt{d}})f_v(\mathbf{H}).
\end{align}
Here, $\mathbf{H}$ is the hidden features, $f_q(\cdot)$, $f_k(\cdot)$, and $f_v(\cdot)$ are the learnable projection functions to calculate the query, key, and values.  

In a transformer, the attention mechanism operates through multiple self-attention layers. Each layer consists of multiple attention heads, which learn different representations and capture diverse aspects of the input sequence. By employing self-attention across multiple layers, transformers can effectively model both local and global dependencies in the input sequence. 

Furthermore, LLMs are typically pre-trained on massive amounts of text data using unsupervised learning objectives. The most commonly used task is language modeling, which is formulated as:
\begin{align}
\mathcal{L}=-\log p(s_i|s_{1:i-1}),
\end{align}
where $s_{1:i}$ is a natural language sequence.
This pre-training phase enables the models to acquire a rich understanding of language patterns and structures, and commonsense knowledge. The pre-trained LLMs can then be fine-tuned on specific downstream tasks with task-specific supervised learning, allowing them to adapt their knowledge to specific tasks and domains.

\section{Method}\label{sec:method}
\subsection{Overall Framework}
We aim to leverage LLMs to provide predictions for TAG tasks while enabling interpretability. The main challenge lies in efficiently equipping the LLM with neighborhood knowledge of the target nodes in the TAG. To address this, our framework incorporates disentangled graph learning, which allows us to learn rich semantic information from the neighborhood and inject it into the downstream LLM. An overall framework of our method is shown in Figure~\ref{fig:framework}. Next, we elaborate our proposed method.
\begin{itemize}[leftmargin = *]
    \item In \textbf{Step 1}, we generate text embeddings by computing the average of the features at the last layer in the upstream LLM. This process captures the overall context and semantics of the text associated with each node in the TAG, which form the foundation for subsequent steps. 
    \item In \textbf{Step 2}, we employ disentangled graph learning to learn embeddings that incorporate diverse neighborhood information. The disentangled graph learning approach allows us to capture varied information from the neighborhood, facilitating a comprehensive understanding of the TAG.
    \item In \textbf{Step 3}, we inject the learned features with neighborhood information into the downstream LLM. This integration of the LLM and the disentangled GNN facilitates more accurate and informed predictions for the downstream TAG tasks.
\end{itemize}
The combination of disentangled graph learning and information injection in the downstream LLM forms the core of our framework. This approach allows us to harness the strengths of both graph representation learning and language modeling, enabling effective utilization of structural information and enhancing the interpretability of our predictions. 

Moreover, we adopt a scheme where the pre-trained LLMs remain frozen in our approach. We only focus on tuning the disentangled GNN to generate structural information for LLMs. This scheme benefits us from several advantages. Firstly, we keep a very low computation cost on fine-tuning since the disentangled GNN only cover a small rate of the parameters.  Even if the LLM is updated to a new version, we can quickly adapt to the new one with minimal tuning procedure. Secondly, our approach maximizes the utilization of LLMs' pre-existing knowledge sand mitigates the risk of catastrophic forgetting.

\begin{algorithm}[t]
\caption{The training process of DGTL} 
\label{alg:dgtl}
\begin{algorithmic}[1]
\REQUIRE Text-attributed graph $\mathcal{G}=(\mathbf{T},\mathbf{A})$,  LLM $L_{\text{up}},L_{\text{down}}$
\ENSURE Parameters of the disentangled GNN $\theta$
\STATE Get node embedding $\mathbf{H}^{(0)}$ of $\mathbf{T}$ by $L_{\text{up}}$\;
\STATE Initialize $\theta$ randomly\;
\WHILE{\text{not converge}}
    \STATE Sample a mini-batch from the training set
    \STATE Compute $\mathbf{H}_i^{(1)} = \text{GNN}_{i,1}(\mathbf{H}^{(0)},\mathbf{A})$
    \STATE Compute disentangled graph structures $\mathbf{A}_i$ by Eq.~\eqref{equ:disen}
    \STATE Compute $\mathbf{H}_i^{(2)} = \text{GNN}_{i,2}(\mathbf{H}_i^{(1)},\mathbf{A}_i)$
    \FOR{node $u$ in the mini-batch}    
    \STATE Construct prompt and response template of $u$
    \STATE Inject $\mathbf{h}_u^{(2)}$ into $L_{\text{down}}$ by Eq.~\eqref{equ:inj1}\eqref{equ:inj2}\eqref{equ:inj} during generation
    \STATE Generate prediction and calculate the loss function
    \ENDFOR
    \STATE Update $\theta$ by gradient descent
\ENDWHILE
\end{algorithmic}
\end{algorithm}

\subsection{Text Embedding Generation}
The first step in our method is to generate an effective text embedding that encapsulates the semantics and contextual information of the input text. Traditionally, previous works\cite{chen2023exploring} have primarily focused on utilizing the embedding of the End-Of-Sentence (EOS) token at the last layer as the text embedding. However, we propose that taking the average of the hidden states in the last layer provides a more comprehensive representation of the entire input text.

Considering the average of the hidden states enables us to capture the collective information and contextual understanding of the text across all positions in the input sequence. This approach allows us to incorporate a broader range of linguistic cues and dependencies into the text embedding, resulting in a more robust and informative representation.


\subsection{Disentangled Graph Learning}
Our next step is learning hidden features that capture relevant neighborhood information for the target task. By injecting these features into the downstream LLM, we enable the LLM to effectively utilize the complex neighborhood information of the nodes in the TAG to assist in predicting the downstream task.

To achieve this, we employ disentangled graph learning, which allows us to capture and represent the diverse neighborhood information present in the TAG. Specifically, we adopt multiple parallel 2-layer GNNs to learn the features, and disentangle them by assigning diverse graph structures. We use a parameter-efficient and differentiable manner to generate the weights of edges for the graph structures, which can be formulated as follows:
\begin{align}
\label{equ:disen}
\mathbf{A}_{i,(u,v)} = \delta \mathbf{A}_{(u,v)}+(1-\delta)\mathbf{A}_{(u,v)}\cdot \sigma((\mathbf{S}_i^u \mathbf{h}_u)^\top(\mathbf{S}_i^v \mathbf{h}_v)),
\end{align}
where $\mathbf{A}_{i,(u,v)}$ is the weight of edge between node $u$ and $v$ in the adjacent matrix of the $i$-th GNN, $\mathbf{A}_{(u,v)}$ indicates if there is an edge between node $u$ and $v$ in the original graph, $\mathbf{S}_i^u$ and $\mathbf{S}_i^v$ are learnable parameters to generate edge weights of the $i$-th GNN, $\delta$ is the hyper-parameter, and $\sigma(\cdot)$ denotes the sigmoid function. Consequently, we generate diverse graph structure by assign different edge weights in a continuous space. We can also conveniently use gradient based methods to optimize the parameters.

Our disentangled GNN architecture incorporates diverse graph structures to ensure the learning of varied information from the neighborhood. These graph structures highlight different aspects of the TAG's topology and semantics, enabling the following LLM to leverage specific types of neighborhood information during downstream task predicting.


\begin{table}
\centering
\caption{Dataset Statistics.}
\vspace{5pt}
\label{tab:dataset}
\begin{center}
\begin{small}
\scriptsize
\begin{tabular}{lrrrl}
\toprule
Dataset & Nodes & Edges & Classes & Domain\\
\midrule
Cora & 2,708 & 5,429 & 7 & Academic\\
PubMed & 19,717 & 44,338 & 3 & Academic\\
Books-History & 41,551 & 358,574 & 12 & E-commerce\\
\bottomrule
\end{tabular}
\end{small}
\end{center}
\end{table}

\subsection{Neighborhood Information Injection}
To take full advantages of the LLMs' ability to understand and model complex patterns and semantics in TAGs, we inject the learned disentangled embedding directly into the downstream LLM.

Specifically, we reserved a set of token positions for placing these disentangled embeddings in the prompt input to the downstream LLM. Even if these embeddings are not in the form of natural language, the form of our input will still allow the LLM to think that they are aligned with the natural semantic space that humans can understand. In this way, we can optimize our disentangled GNNs by gradient descent methods during fine-tuning.

However, the gradient back-propagation through the entire LLM neural architecture can make the optimization process extremely difficult during the fine-tuning procedure. Therefore, we take a step further by performing the embedding injection in all layers of the downstream LLM. Specifically, we add the disentangled embedding to the text embedding at the reserved positions in key and query projection functions as follows:
\begin{align}
\label{equ:inj1}
f_{\{q,k\}}(\mathbf{x}_i) = \mathbf{W}_{\{q,k\}}(\mathbf{x}_i+\mathbf{p}_{\{q,k\}}^{(i)}+\mathbf{h}_u^{(i)}),
\end{align}
where $\mathbf{h}_u^{(i)}$ is the disentangled embedding of node $u$ placed in position $i$, $\mathbf{W}_{\{q,k\}}$ is the projection matrix, $\mathbf{x}_i$ is the corresponding feature from the last layer in the LLM, and $\mathbf{p}_{\{q,k\}}^{(i)}$ indicates the absolute or relative position embedding of position $i$. As an alternative, we use the following function for rotary position embedding:
\begin{align}
\label{equ:inj2}
f_{\{q,k\}}(\mathbf{x}_i) = \mathbf{R}_i\mathbf{W}_{\{q,k\}}(\mathbf{x}_i+\mathbf{h}_u^{(i)}),
\end{align}
where $\mathbf{R}_i$ is the rotary matrix of the position embedding. Due to the varying position encoding of the injected features, our disentangled GNNs are encouraged to learn diverse knowledge, further enriching the semantic knowledge learned by different parts of GNNs. In addition, we also use the disentangled embedding to in value calculation by a similar way:
\begin{align}
\label{equ:inj}
f_v(\mathbf{x}_i) = \mathbf{W}_v(\mathbf{x}_i+\mathbf{h}_u^{(i)}).
\end{align}
As such, our method incorporates neighborhood information from GNNs into each layer of the large language model LLM, enabling the LLM to benefit from a comprehensive understanding of the graph structure throughout the entire model. This injection of information in all layers facilitates a direct gradient flow to the GNNs, resulting in more accurate and informative gradient updates. This integration of language modeling and graph representation learning allows our model to leverage the contextual information captured by the LLM and the structural patterns learned by the GNNs, leading to effective learning and improved performance.

\subsection{Fine-tuning Procedure}

\begin{table*}
  \caption{The illustration of prompts and typical responses of different types of datasets. The \textcolor{blue!80!white}{blue} parts indicate the texts populated based on the content of the dataset, the \textcolor{red!80!black}{red} parts indicate the embeddings obtained by fine-tuning based on neighborhood information. and the \textcolor{green!60!black}{green} parts indicate the prediction given by the LLM.}
  \label{tab:prompt}
  \vspace{5pt}
  \small
  \begin{tabularx}{\textwidth}{
  >{\hsize=0.25\hsize\linewidth=\hsize}X|
  >{\hsize=0.35\hsize\linewidth=\hsize}X|
  >{\hsize=1.4\hsize\linewidth=\hsize}X
}
    \toprule
    \multicolumn{2}{l|}{Dataset} & Content\\
    \midrule
    \multirow{2}{*}{Citation Dataset} & Prompt & Here is a paper title and abstract: \textcolor{blue!80!white}{(paper content)}. Some information about the references cited in this paper: \textcolor{red!80!black}{(neighbor content)}. Task: there are following categories: \textcolor{blue!80!white}{(list of categories)}. Which category does this paper belong to? Output the most 1 possible category of this paper.\\\cline{2-3}
    & Response Template& Based on the content of the paper, the most appropriate category for this paper would be \textcolor{green!60!black}{(category prediction)}. $\cdots$\\
    \midrule
    \multirow{2}{*}{\tabincell{l}{E-commerce \\ Dataset} } & Prompt & Here is a book description and title: \textcolor{blue!80!white}{(book content)}. Some information about the books frequently purchased together: \textcolor{red!80!black}{(neighbor content)}. Task: there are following categories: \textcolor{blue!80!white}{(list of categories)}. Which category does this book belong to? Output the most 1 possible category of this book.\\\cline{2-3}
    & Response Template& Based on the information of the book, the most appropriate category for this book would be \textcolor{green!60!black}{(category prediction)}. $\cdots$\\
    \bottomrule
\end{tabularx}
\end{table*}

We summarize the fine-tuning procedure in Algorithm.~\ref{alg:dgtl}. After generating node embedding by LLMs for the input graph at the beginning of the algorithm, the algorithm iteratively updates the parameters until convergence. Each iteration involves sampling a mini-batch from the training set and performing forward propagation through the disentangled GNN layers to capture both the graph structure and the text attributes. We follow the classic auto-regressive scheme to fine-tune. For each node in the mini-batch, we construct its prompt (refer to Section~\ref{sec:prompt} for more details) as the input. Since we are only concerned about the LLM's prediction for node categories, we design a response template (some examples are shown in Table~\ref{tab:prompt}) as part of the prompt. Consequently, the LLM can directly predict the category at the first token of the generation procedure. Then we can conveniently calculate the loss function with respect to the category prediction, which can be formulated as follows:
\begin{align}
\mathcal{L}=-\log p(s_{\text{lab}}|s_{\text{pro}}+s_{\text{res}}),
\end{align}
where $s_{\text{lab}}$ indicates the label token, $s_{\text{pro}}$ and $s_{\text{res}}$ indicate the sequences of the prompt and response template. We find that only calculating the loss function of the first token can work well in practice. As such, we can fine-tune the model in a more targeted manner, enabling it to learn the most useful information for downstream classification tasks. Back-propagation is performed after each node classification, and the parameters $\theta$ are updated through a gradient descent method when all nodes in the mini-batch are predicted. Through this iterative process, DGTL learns a text generation model that leverages the rich information present in text-attributed graphs.

\section{Experiments}\label{sec:exp}

\subsection{Experimental Setup}
\subsubsection{Datasets}

We conducted experiments on benchmark TAGs to assess the effectiveness of our framework. These TAGs encompass both citation networks and e-commerce networks, providing a diverse range of domains for evaluation.

For the citation networks, we utilize widely recognized datasets including Cora and PubMed~\cite{sen2008collective}. The Cora dataset focuses on computer science research papers. 
PubMed, on the other hand, is specific to biomedical literature, making it particularly relevant for medical-related tasks. These datasets consist of academic papers as nodes and their citation relationships as edges. Each paper has its title and abstract as text information. The task on these citation networks is to determine the category of the papers.

In addition, we also incorporated e-commerce graphs into our evaluation. Specifically, we employ Book-History dataset~\cite{TAG-Benchmark}, which contains a rich variety of historical books. In the dataset, nodes represent books, while the edges indicate the connected books are frequently purchased together. Each book also has its description as the text information. The task on the e-commerce network is to determine the category of the books.

The statistics of the datasets are shown in Table~\ref{tab:dataset}. For all datasets, we follow the classical semi-supervised learning setting, i.e., we random select 20 nodes of each category as the training set, and randomly select 1,000 nodes from the rest nodes as the test set. 

\subsubsection{Implementations}\label{sec:prompt}

\begin{table*}
\vspace{-5pt}
\caption{Performance comparison with GNN predictor baselines.}
\label{tab:main}
\vspace{5pt}
\begin{center}
\small
\begin{tabular}{l|ccc|ccc}
\toprule
Dataset & \multicolumn{3}{c|}{Cora} & \multicolumn{3}{c}{PubMed} \\\midrule
GNN backbone & GCN & GAT & MLP & GCN & GAT & MLP  \\
\midrule
TF-IDF & $81.99_{\pm 0.63}$ & $82.30_{\pm 0.65}$ & $67.18_{\pm 1.01}$ & $78.86_{\pm 2.00}$ & $77.65_{\pm 0.91}$ & $71.07_{\pm 0.78}$\\
Word2Vec & $74.01_{\pm 1.24}$ & $72.32_{\pm 0.17}$ & $55.34_{\pm 1.31}$ & $70.10_{\pm 1.80}$ & $69.30_{\pm 0.66}$ & $63.48_{\pm 0.54}$\\
Deberta-base & $48.49_{\pm 1.86}$ & $51.02_{\pm 1.22}$ & $30.40_{\pm 0.57}$ & $62.08_{\pm 0.06}$ & $62.63_{\pm 0.27}$ & $53.50_{\pm 0.43}$\\
Fine-tuned Deberta-base & $59.23_{\pm 1.16}$ & $57.38_{\pm 2.01}$ & $30.98_{\pm 0.68}$ & $62.12_{\pm 0.07}$ & $61.57_{\pm 0.07}$ & $53.65_{\pm 0.26}$\\
Llama-7B & $66.80_{\pm 2.20}$ & $59.74_{\pm 1.53}$ & $52.88_{\pm 1.96}$ & $73.53_{\pm 0.06}$ & $67.52_{\pm 0.07}$ & $66.07_{\pm 0.56}$\\
Sentence-BERT & $82.20_{\pm 0.49}$ & $82.77_{\pm 0.59}$ & $74.26_{\pm 1.44}$ & $81.01_{\pm 1.32}$ & $79.08_{\pm 0.07}$ & $76.66_{\pm 0.50}$\\
e5-large & $82.56_{\pm 0.73}$ & $81.62_{\pm 1.09}$ & $74.26_{\pm 0.93}$ & $82.63_{\pm 1.13}$ & $79.67_{\pm 0.80}$ & $80.38_{\pm 1.94}$\\
GLEM-GNN & $48.49_{\pm 1.86}$ & $51.02_{\pm 1.22}$ & - & $62.08_{\pm 0.06}$ & $62.63_{\pm 0.27}$ & -\\
GLEM-LM & $59.23_{\pm 1.16}$ & $57.38_{\pm 2.01}$ & - & $62.12_{\pm 0.07}$ & $61.57_{\pm 0.07}$ & -\\
\midrule
DGTL & $81.10_{\pm 0.20}$ & $80.40_{\pm 0.42}$ & - & $87.10_{\pm 1.31}$ & $85.63_{\pm 1.54}$ & -\\
\bottomrule
\end{tabular}
\end{center}
\end{table*}

\begin{table}
\vspace{-5pt}
\caption{Performance comparison with LLM predictor baselines and the ablation model.}
\label{tab:llm}
\begin{center}
\scriptsize
\begin{tabular}{l|ccc}
\toprule
Method & Cora & PubMed & Books-History\\
\midrule
0-hop & $60.80$ & $62.30$ & $17.70$\\
Neighbor Summary & $77.30$ & $67.70$ & $17.90$ \\
\midrule
DGTL & $81.10_{\pm 0.20}$ & $87.10_{\pm 1.54}$ & $55.70_{\pm 0.89}$\\
\midrule
DGTL w/o disen & $80.67_{\pm 0.47}$ & $85.00_{\pm 1.47}$ & $55.40_{\pm 1.04}$ \\
\bottomrule
\end{tabular}
\end{center}
\end{table}

In our experiments, we employed Llama-2-13B-chat~\cite{llama2} as the backbone LLM. Llama2 is a representative LLM known for its impressive language modeling capabilities and extensive pre-training on a vast corpus of text data. By utilizing Llama2 as our backbone LLM, we aimed to validate the effectiveness of our approach when collaborated with LLMs.

To tailor our framework to different datasets and tasks, we designed specific prompts for different types of datasets. The prompts served as initial instructions or cues provided to the LLM to guide our fine-tuning for the target task. 
The details of the designed prompts are summarized in Table~\ref{tab:prompt}. The table outlines the specific prompt structure, including any additional instructions or formatting applied, for each dataset used in our experiments. By customizing the prompts, we ensured that the LLM could effectively adapt its pre-trained knowledge to the specific requirements and characteristics of each dataset.

For other hyper-parameters, we set the number of disentangled channels is 16 for citation datasets, and 8 for the e-commerce datasets. The number of hidden dimensions of each disentangled GNN channel is 32. The hyper-parameter to control the disentangled structure $\delta$ is 0.8. For optimization, we use the Adam optimizer with a learning rate of 0.001.

\subsubsection{Baselines}
We compare our model with baselines from the following two categories.
\begin{itemize}[leftmargin = *]
    \setlength{\parskip}{0pt}
    \item \textbf{GNN Predictors}. Following Chen at el.~\cite{chen2023exploring}, we consider different language models to enhance the node features of the TAG, including Deberta~\cite{deberta}, Llama~\cite{llama}, Sentence-BERT~\cite{sentencebert}, and e5~\cite{e5large}. Shallow embeddings TF-IDF and Word2vec are also included. We include GLEM~\cite{zhao2022learning} as a baseline as well. For all these feature enhancer methods, three GNN backbones, including GCN~\cite{gcn}, GAT~\cite{gat}, and MLP are adopted. For these baselines, we directly use the classification accuracy as the evaluation metric.
    \item \textbf{LLM predictors}.
    We also consider using LLM (Llama2-13B-chat) as the predictors as our baselines, including using prompts without neighborhood information (0-hop prompt) and neighborhood summarization~\cite{chen2023exploring}. For these methods, we use exact match scores~\cite{rajpurkar2016squad}, i.e., the percentage of predictions that match any one of the ground truth, as the metric.
truth answers exactly.
\end{itemize}

\subsection{Experimental Results}

\begin{table*}
  \caption{The interpretation comparison with LLM predictor methods on two cases. Case 1: a paper classification task on Cora dataset. Case 2: a book classification task on Book-History dataset. The \textcolor{blue!80!white}{blue} parts are the labels. The \textcolor{red!80!black}{red} parts are the wrong predictions, while the \textcolor{green!60!black}{green} parts are the correct predictions and the corresponding structurally relevant interpretations.}
  \label{tab:explain}
  \vspace{5pt}
  \small
  \begin{tabularx}{\textwidth}{
  >{\hsize=0.4\hsize\linewidth=\hsize}X|
  >{\hsize=0.5\hsize\linewidth=\hsize}X|
  >{\hsize=0.5\hsize\linewidth=\hsize}X|
  >{\hsize=0.6\hsize\linewidth=\hsize}X
}
    \toprule
    Text & 0-hop & Neighbor Summary & DGTL\\
    \midrule
    \textcolor{blue!80!white}{(label: rule learning)} Here is a paper title and abstract:  Limited Dual Path Execution. : This work presents a hybrid branch predictor scheme that uses a limited form of dual path execution along with dynamic branch prediction to improve execution times.$\cdots$  & Based on the content of the paper, the most appropriate category for this paper would be "\textcolor{red!80!black}{theoretical}" or "\textcolor{red!80!black}{probabilistic methods}". The paper presents a new approach to branch prediction that uses a hybrid predictor model that combines dynamic branch prediction with limited dual path execution.$\cdots$  &  Based on the content of the paper, the most appropriate category for this paper would be '\textcolor{red!80!black}{theory}'. The paper presents a new approach to branch prediction and speculative execution, and proposes a hybrid predictor model that combines dynamic branch prediction and dual path execution. $\cdots$ &  Based on the content of the paper, the most appropriate category for this paper would be '\textcolor{green!60!black}{rule learning}'. The paper is discussing the use of a hybrid branch predictor scheme that uses a limited form of dual path execution along with dynamic branch prediction to improve execution times, which is a type of rule-based approach. \textcolor{green!60!black}{The paper also cites several references related to rule-based machine learning}, $\cdots$\\
    \midrule
    \textcolor{blue!80!white}{(label:military)} Here is a book description and title: Description: Between Frederick Buechners profound autobiography Sacred Journey and Russell Bakers warm and humorous Growing Up, Matt Matthews One Thousand Miles charts a compelling journey all its own through the past and the present.$\cdots$ & Based on the information provided, the book "One Thousand Miles" by Matt Matthews belongs to the category of "\textcolor{red!80!black}{Memoir/Biography}" or "\textcolor{red!80!black}{History}". The book is described as a "compelling journey through the past and present" that explores themes of identity, family, and the human experience, and draws on a wide range of sources including autobiography, history, and culture. $\cdots$ & Based on the book description, the most appropriate category for "Matt Matthews One Thousand Miles" would be "\textcolor{red!80!black}{Americas}" or "\textcolor{red!80!black}{North American Historical Study.}" The book is described as a story of a son's pilgrimage into the heart of his father's story, set against the backdrop of World War II and the generation that endured it. $\cdots$ & Based on the information of the book, the most appropriate category for this book would be '\textcolor{green!60!black}{Military History}'. The book is about a son's journey to understand his father's experience during WWII, which is a historical event. \textcolor{green!60!black}{There are also other books that are frequently purchased together, which are related to military history, such as 'The Second World War' and 'The Cold War'.} Therefore, the most appropriate category for this book would be 'Military History'.\\
    \bottomrule
\end{tabularx}
\end{table*}


The result comparison with the GNN predictor baselines is shown in Table~\ref{tab:main}. From the table, we can find that GNN backbones are generally better than MLP backbone, indicating that structural information is essential to solve TAG tasks. Our method achieves comparable performance to SOTA baselines on Cora and outperforms all baselines on PubMed. Besides, our method offers a distinct advantage in terms of interpretability. Unlike the baselines, our method can provide natural language explanations for the predictions, enhancing the transparency and comprehensibility of the model's decision-making process. This interpretability aspect is particularly valuable in scenarios where understanding the reasoning behind the predictions is crucial. 

The result comparison with the LLM predictor baselines is shown in Table~\ref{tab:llm}. Although neighbor summary prompt achieves better performance than 0-hop prompt by considering neighborhood information, our method outperforms these baselines by a large margin, indicating that our method enables LLMs to effectively learn and utilize task-relevant knowledge and benefits the downstream task prediction.

\subsection{Interpretability with Structural Information}

We showcase some examples to illustrate the interpretation of our method. The results as well as the comparison with two LLM predictor baselines are shown in Table~\ref{tab:explain}.

The results indicate that our proposed method is more capable of predicting the correct label on the target node while the other methods fail. Moreover, our method also generates explanations related to neighborhood information after giving the prediction. For example, in the first case, our method predicts the paper belongs to "rule learning", and a supporting evidence for this prediction is "the  paper also cites several references related to rule-based machine learning." This supporting evidence is exactly derived from the neighborhood information on the TAG, which demonstrates that our method can effectively capture the semantic information on the graph to make more accurate predictions. In case 2, our method predicts that the book belongs to military history. In addition to making predictions based on the content of the book, it also utilize the neighborhood information in the TAG to assist in the prediction process, as it states "there are also other books that are frequently purchased together, which are related to military history, such as 'The Second World War' and 'The Cold War'".

The shown cases demonstrate that our method generates interpretations for classification predictions that are accurate and structurally relevant. By contrast, the baselines produce incorrect predictions without using structural information for interpretation. Our method effectively integrates the graph's structural information in LLM generation, providing meaningful insights and justifications for the classification results. The experimental results highlight the importance of incorporating structural information in achieving accurate and interpretable predictions in TAG tasks. More importantly, through our approach, humans can harness intuitive interpretations based on graph structures when using LLMs to tackle TAG problems. This greatly enhances the utilization of knowledge embedded in TAG and unleashes the potential of LLMs on graphs.

\subsection{Ablation Study}

In this section, we evaluate the effectiveness of the disentanglement component of our method through an ablation study. The results are also shown in Table~\ref{tab:llm}. The results demonstrate that the disentanglement is greatly beneficial and crucial for our method to learn better structural information and enables the downstream LLM to give more accurate predictions accordingly.

\section{Conclusion}\label{sec:con}
In conclusion, this paper addresses the challenge of effectively integrating structural information into large language models (LLMs) for text-attributed graph (TAG) tasks. We propose the Disentangled Graph Text Learner (DGTL) model, which leverages tailored disentangled graph neural network layers to capture complex graph structure information in TAGs. Our method enhances the reasoning and prediction capabilities of LLMs while providing natural language explanations for model predictions, which is crucial for interpretability in TAG tasks. Through extensive experiments on various TAG benchmarks, we demonstrate that DGTL achieves competitive performance compared to state-of-the-art baselines while offering human-understandable explanations for model predictions. Our work contributes to advancing the field of TAG analysis by harnessing the power of LLMs and improving their interpretability for real-world applications.

\bibliography{mypaper}
\bibliographystyle{icml2023}



\end{document}